\documentclass{article}
\usepackage{ijcai19}

\usepackage{times}
\usepackage{soul}
\usepackage{url}
\usepackage[hidelinks]{hyperref}
\usepackage[utf8]{inputenc}
\usepackage[small]{caption}
\usepackage{graphicx}
\usepackage{amsmath}
\usepackage{booktabs}
\usepackage{algorithm}
\usepackage{algorithmic}
\usepackage{makecell}
\usepackage{bm}
\usepackage{booktabs} 

\usepackage{multirow}
\usepackage[pagewise]{lineno}
\usepackage{comment}
\usepackage{color, colortbl}
\definecolor{Gray}{gray}{0.92}

\title{Travel Time Estimation without Road Networks: An Urban Morphological Layout Representation Approach}

\author{
Wuwei Lan$^{1,}$\thanks{Equal Contribution}
\and
Yanyan Xu$^{2,*,}$\thanks{Contact Author}
\and Bin Zhao$^3$
\affiliations
$^1$Department of Computer Science and Engineering, Ohio State University, Columbus, OH 43210\\
$^2$Department of City and Regional Planning, University of California, Berkeley, CA 94720\\
$^3$Wisense AI, Jinan, China\\
\emails
lan.105@osu.edu,
yanyanxu@berkeley.edu,
binzhao@powergrid.ai
}

\begin{document}

\maketitle

\begin{abstract}
Travel time estimation is a crucial task for not only personal travel scheduling but also city planning. Previous methods focus on modeling toward road segments or sub-paths, then summing up for a final prediction, which have been recently replaced by deep neural models with end-to-end training. Usually, these methods are based on explicit feature representations, including spatio-temporal features, traffic states, etc. Here, we argue that the local traffic condition is closely tied up with the land-use and built environment, i.e., metro stations, arterial roads, intersections, commercial area, residential area, and etc, yet the relation is time-varying and too complicated to model explicitly and efficiently. Thus, this paper proposes an end-to-end multi-task deep neural model, named \textit{Deep Image to Time} (DeepI2T), to learn the travel time mainly from the built environment images, a.k.a. the morphological layout images, and showoff the new state-of-the-art performance on real-world datasets in two cities. Moreover, our model is designed to tackle both path-aware and path-blind scenarios in the testing phase. 
This work opens up new opportunities of using the publicly available morphological layout images as considerable information in multiple geography-related smart city applications.
\end{abstract}

\section{Introduction}
Travel time estimation in the urban area is vital to individual travel planning, transportation and city planning. Timely estimation of travel time help travelers to effectively schedule their trips in advance, plan the charging of electric vehicles~\cite{xu2018planning}, evaluate travel exposure to air pollution~\cite{xu2019unraveling}, and help transportation network companies to improve the service quality of delivery vehicles~\cite{mori2015review}. From transportation planning perspectives, travel time estimation can facilitate the quantification of individual driver's contribution to the overall traffic congestion~\cite{ccolak2016understanding,xu2017collective}. Travel time is also one of the most important metrics to evaluate residents' accessibility to resources in city planning~\cite{weiss2018global}. However, travel time estimation in traffic is still challenging due to the complexity of transportation systems and the unpredictability of individual travel needs and mobility behavior, especially in urban regions.

This work places the emphasis on \textit{travel time estimation} for a trip query in urban environment utilizing massive trajectory data. The developed model desires to tackle not only the \textit{path-aware} query, where the routing path is available, but also the \textit{path-blind} query, which provides the origin and destination locations only.
Recent solutions are proposed in two aspects (i) link-based and (ii) path-based approaches. The former first individually model the travel time on each traversed link and then accumulate them for a given path. The main drawbacks of these approaches are the accumulation of error and the ignorance of travel delay at the intersections and traffic signals. Besides, they can not directly work for the \textit{path-blind} travel time estimation~\cite{woodard2017predicting}. Path-based approaches aim to directly estimate the travel time of the whole path. There are two ways to preprocess the path before training models, mapping the path to road networks via map-matching~\cite{li2018multi}, which is computationally expensive for massive trajectory data, and to grid cells~\cite{zhang2018deeptravel}. Regarding the gridding methods, the varying traffic states in one grid is intractable to capture as there might be multiple roads in the same grid and the traffic states on different segments and directions are dramatically divergent.

\begin{figure}[tb!]
\centering
\centerline{\includegraphics[width=0.75\linewidth]{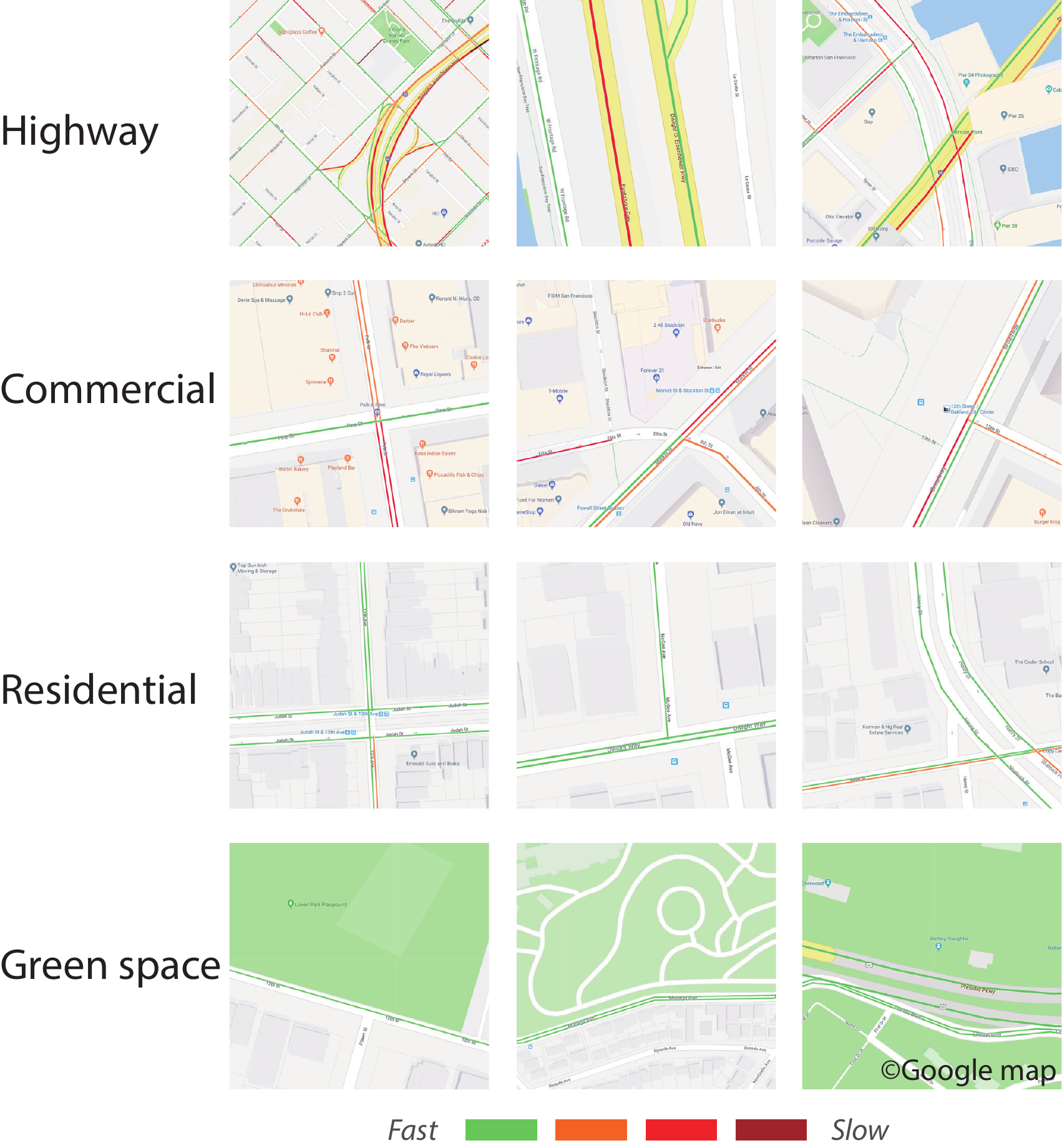}}
\caption{Illustrations of different morphological layouts with traffic states in urban environment, provided by Google Maps.}
\label{fig:layout}
\end{figure}

Inspired by the relation between traffic congestion and urban land use and organization~\cite{tsekeris2013city,louf2013modeling,lee2017morphology}, we desire to capture the congestion level of local regions from their morphological layouts. Figure~\ref{fig:layout} illustrates diverse morphological layouts in an urban area with additional traffic states in Google Maps. The layouts provide rich and learnable information about the built environments, including transportation infrastructure (levels of roads are differentiated by colors or widths), green spaces, density of buildings, commercial regions, etc.~\cite{albert2017using}. The variant built environments imply the nontrivial yet easy to be neglected connection between traffic congestion and the layout. Thus, appropriate representation of layout images could be a significant proxy of traffic states. The travel delay would be heavy if a driver traverses busy regions, such as the regions with dense traffic facilities (e.g., metro stations) or commercial facilities.

Taking cognizance of the relation between built environment and traffic congestion, in this paper, we are interested in the question "\textit{Could we learn the travel delay from the urban layout images?}" To this end, we present an end-to-end multi-task deep learning model, named \textit{\ul{Deep} \ul{I}mage \ul{to} \ul{T}ime} (DeepI2T), to estimate the travel time of a path with the representation of layout images of the traversed grid sequence. Our main contributions are summarized as follows.

\begin{itemize}
    \item We propose an end-to-end multi-task deep learning approach for \textit{travel time estimation} by integrating the trajectory data with morphological layout images. To the best of our knowledge, this is the first time to introduce the fine-scale layout images in transportation.
    \item DeepI2T learns the travel delay during the whole paths and sub-paths from the gridding images, without the use of road networks, hence without map-matching. 
    \item We combine the layout images in grids with driving direction of each vehicle. Heterogeneous traffic conditions in one single grid could be represented distinctively.
    \item DeepI2T could work for both \textit{path-ware} and \textit{path-blind} trip query in the testing stage. A neighboring trips solution is designed to tackle the \textit{path-blind} query.
    \item We showcase DeepI2T with massive trajectory data in two cities. The performance is competitive with several state-of-the-art baselines.
\end{itemize}


\section{Related Work}


According to the information provided by the trip query in the testing phase, these path-based (a.k.a. trajectory-based) approaches fall into two categories, \textit{path-aware} and \textit{path-blind}.

\textit{Path-aware} query provides the specific routing path of the trip to the estimation model. Wang~\textit{et al.} estimated the travel time of each road segment using tensor-based spatial-temporal model, which could handle the roads not traversed by any trajectory~\cite{wang2014travel}. Similarly, Woodard~\textit{et al.} proposed to model the congestion levels on each individual segment using historical trajectory data~\cite{woodard2017predicting}. 
In~\cite{wang2018learning}, the authors formulated the travel time estimation to regression problem and proposed wide-deep-recurrent model feeding with multiple features, including the spatial, temporal, traffic, and personalized features. In these works, for modeling the traffic features on individual road segments, map-matching is a must in the primary stage and the queried trip must provide the taking route to the model (a.k.a., a sequence of road segments).

Thanks to the powerful representation ability of deep neural networks, recent works attempted to directly learn the travel time from trajectory data, without the time-consuming map-matching. 
Zhang~\textit{et al.} first mapped the GPS locations to grids and designed a model to estimate travel time by combining the spatial and temporal embedding with some auxiliary features, including the driving states, short-term and long-term traffic states in grids~\cite{zhang2018deeptravel}. Wang~\textit{et al.} designed an end-to-end framework to learn the spatial and temporal dependencies from the raw GPS sequence~\cite{wang2018will}. During the testing phase, the path of the queries trip is provided as a sequence of GPS locations in a route. As this method is trained on the raw GPS coordinates, its performance is sensitive to the quality of training data and difference between training and testing data.

\textit{Path-blind} query only provides the origin and destination locations and departure time to the estimation model. It's also named as Origin-Destination (OD) travel time estimation and is universal in urban planning for the evaluation of reachability to facilities. In contrast with \textit{path-ware}, \textit{path-blind} query faces with great challenge due to the uncertain route and travel distance. Li~\textit{et at.} built a spatial and temporal graph on the map to learn the prior knowledge from the traces and designed a multi-task framework to learn the path information between origin and destination~\cite{li2018multi}. This work models the road network as an undirected graph, which ignores the divergence of traffic states in different directions.
Although not dealing with the trajectory data, Wang {et al.} proposed a simple baseline for the \textit{path-blind} travel time estimation using only the origin and destination information in the training sets~\cite{wang2016simple}. The idea is to find the neighboring trips for a queries trip and simply scaling their historical travel times. This method can not perform stably when less neighboring trips are available in training sets.




\section{Preliminary}

\begin{figure}[tb!]
\centering
\centerline{\includegraphics[width=0.72\linewidth]{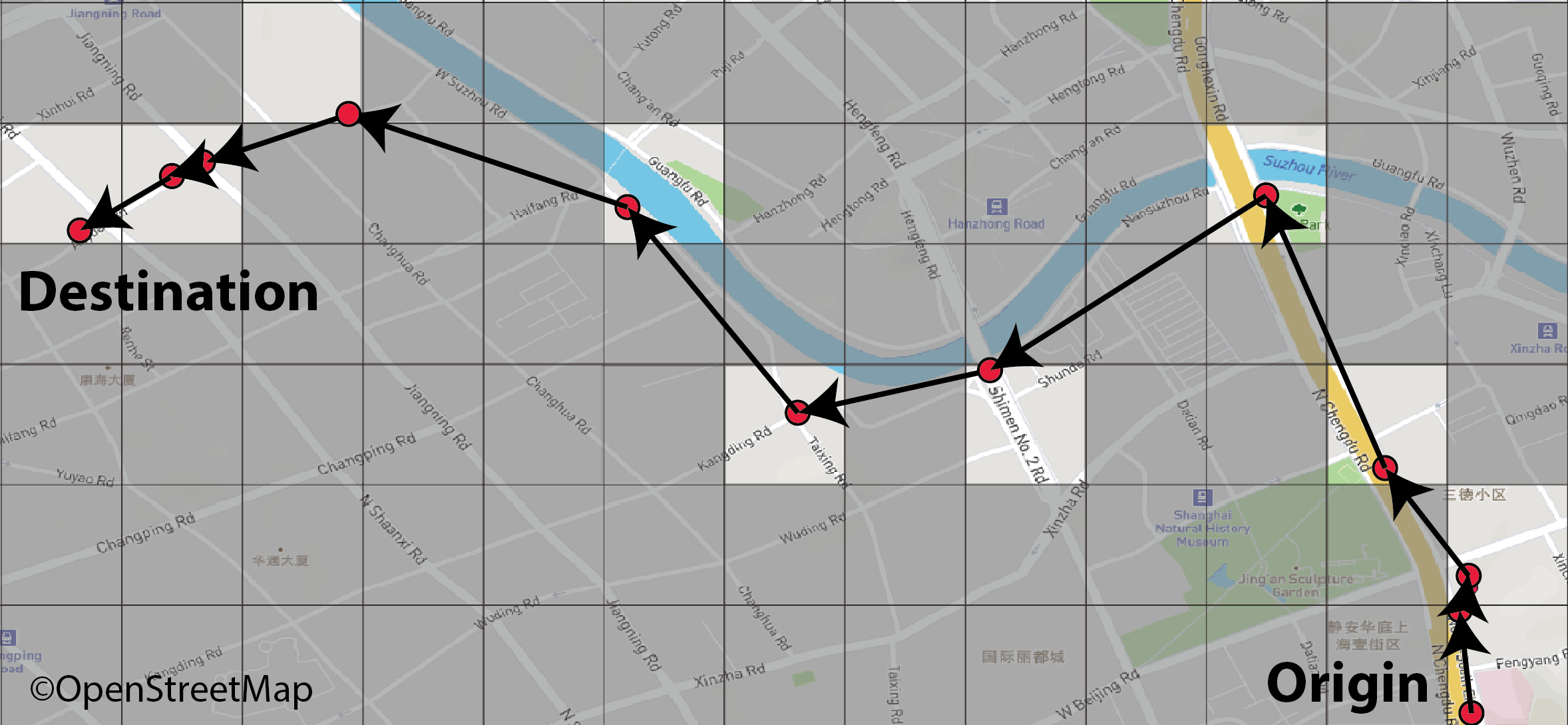}}
\caption{Illustrations of a trajectory and traversed grids.}
\label{fig:traj}
\end{figure}

\paragraph{Driving Trajectory.} The trajectory of a driving trip, $P$, is composed of a sequence of geographical locations $\{Lon, Lat\}$ with timestamps. Each trip is associated with a vehicle ID. Therefore, a trip with $N$ footprints can be formulated as, $P = \{F_1, F_2, ..., F_N\}$, where the $i$th footprint $F_i = (t_i, Lon_i, Lat_i)$. 
The travel time of the path $T_P = t_N - t_1$. The travel distance of the trip equals to the accumulation of great-circle distances between two consecutive footprints, that is, $D = \sum_{i=1}^{N-1} Dist((Lon_i,Lat_i)\rightarrow (Lon_{i+1}, Lat_{i+1}))$. Figure~\ref{fig:traj} illustrates a path with 12 footprints from east to west.

\paragraph{Morphological Layout Images.} The study region is first divided into a number of equal-sized grids. In each grid, we crawl the high-resolutional map from the publicly available map service, OpenStreetMap~\cite{osm2018}, using the Leaflet API~\cite{leaflet2018}. In the morphological layout images, we can visually observe different build environments, i.e., river, park, bridge, expressway, main and secondary roads. As shown in Figure~\ref{fig:traj}, each point in a path could be mapped to a grid image. In this way, the path $P$ could be presented using grid images, $P' = \{(t_1,g_1), (t_2, g_2), ..., (t_N, g_N)\}$, where $g_i$ denotes the traversed grid image at time $t_i$. Note that $g_i$ could be repeated as multiple points might present in one grid.



\section{Model Description.}
\begin{figure}[bt!]
    \centering
    \includegraphics[width=1.0\linewidth]{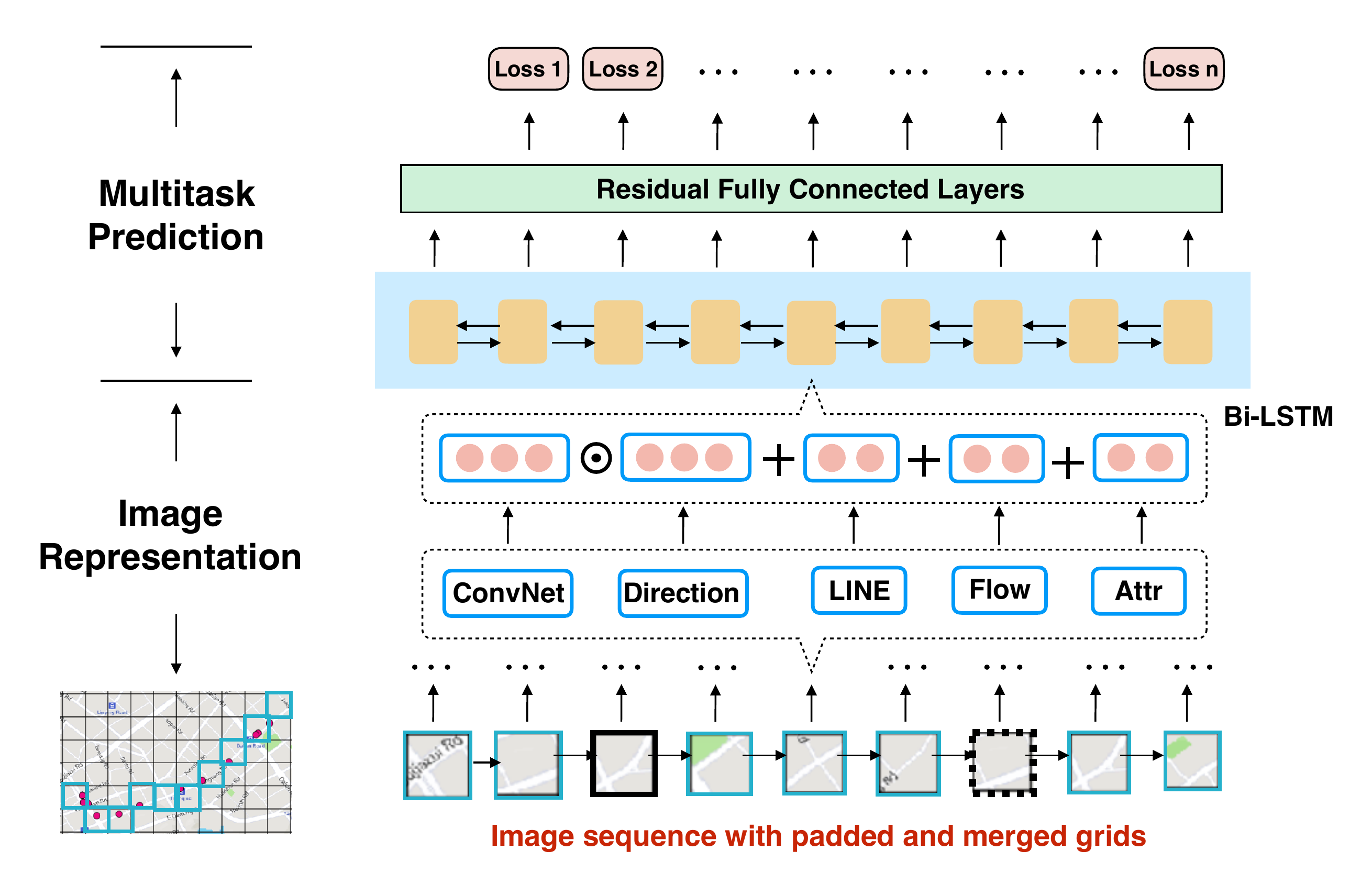}
    \caption{Our proposed DeepI2T architecture. We merge GPS points if they share the same grid (bold frame) and pad new grids (dashed frame) if two continuous GPS points are not neighbors. $\odot$ is element-wise multiplication and $+$ is vector concatenation.}
    \label{fig:model_structure}
\end{figure}

As shown in Figure \ref{fig:model_structure}, our proposed DeepI2T has two components: \textbf{Image Representation} and \textbf{Multitask Prediction}. The first component focuses on extracting feature patterns from morphological layout images using deep convolutional neural networks, while the second component aims at learning sequential dependency among the grids with supervision from multi-task MAPE loss.

\begin{figure}[tbh!]
    \centering
    \includegraphics[width=0.8\linewidth]{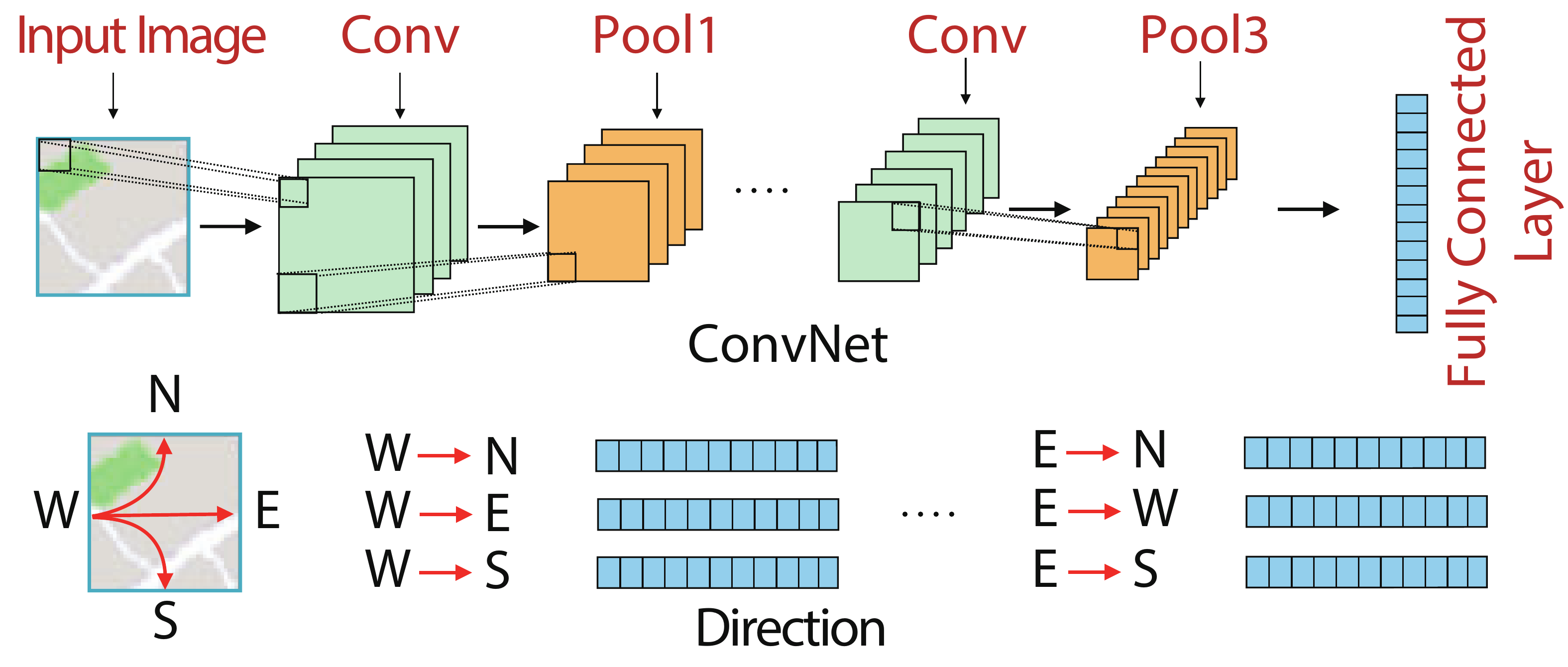}
    \caption{ConvNet architecture and direction embeddings.} 
    \label{fig:ConvNet_structure}
\end{figure}

\subsection{Image Representation}
For each grid image, we apply deep convolutional neural networks to recognize the patterns, which are twisted by direction embedding. We also consider topological information and traffic flow from all the grids, as well as some attribute information, including start time, driver ID and etc. Specifically, we have five modules: ConvNet, Direction, LINE, Flow and Attributes, which we illustrate each as follows:

\paragraph{ConvNet.} Deep convolutional neural networks \cite{krizhevsky2012imagenet} show strong ability in capturing image patterns, e.g. arterial roads, intersections, commercial area and etc. In order to achieve this goal, we designed a 7-layer ConvNet (shown in Figure \ref{fig:ConvNet_structure}), containing 3 convolutional layers, 3 pooling layers, and one fully connected layer for final representation. The detailed parameters of each layer are shown in Table \ref{tab:ConvNet_configuration}. Specifically, we conduct 2-D convolution as
\begin{align}
    y_{i^{l+1}, j^{l+1}, d^{l+1}} =  & \sum_{i=0}^{H}\sum_{j=0}^{W}\sum_{d^{l}=0}^{D^{l}}f_{i,j,d^l, d^{l+1}}*x_{i^{l+1}+i, j^{l+1}+j, d^{l}} \nonumber \\
    & + b_{i^{l+1},j^{l+1},d^{l+1}}  
\end{align}
where \(f_{i,j,d^l, d}\) is an element from kernel vector \(\bf{f}\) with size \((H, W, D^{l}, D^{l+1})\), representing height \(H\), width \(W\), in channels \(D^{l}\) and out channels \(D^{l+1}\). \(x_{i^{l+1}+i, j^{l+1}+j, j, d^{l}}\) refers to an element from \(\bm{x^{l}}\), which is the input data at layer \(l\) with size \((H^{l}, W^{l}, D^{l})\). The output \(y_{i^{l+1}, j^{l+1}, d}\) is an element from \(\bm{y^{l+1}}\) with size \((H^{l+1}, W^{l+1}, D^{l+1})\). The bias term \( b_{i^{l+1},j^{l+1},d^{l+1}}\) is added into \(y_{i^{l+1}, j^{l+1}, d^{l+1}}\). After convolution, we apply Relu function \(Relu(x)=max(0, x)\) for non-linear mapping, and max pooling for down sampling. The last fully connected layer finally outputs 200-dimension vector.

\begin{table}[bt!]
\begin{center}
\footnotesize
\begin{tabular}{lcccc}
 \hline
 & \textbf{Type} & \textbf{Kernel size} & \textbf{Stride} & \textbf{Input size}\\
 \hline
Layer 1 & Conv2D & \(3*3\) & \(1\) & \(3*436*373\) \\
Layer 2 & Pooling & \(2*2\) & \(2\) & \(8*436*373\) \\
Layer 3 & Conv2D & \(3*3\) & \(1\) & \(8*218*187\) \\
Layer 4 & Pooling & \(3*3\) & \(3\) & \(16*218*187\) \\
Layer 5 & Conv2D & \(3*3\) & \(1\) & \(16*73*63\) \\
Layer 6 & Pooling & \(3*3\) & \(3\) & \(8*73*63\) \\
Layer 7 & Full & \(3*3\) & \(1\) & \(8*25*21\) \\
 \hline
\end{tabular}
\end{center}
\caption{Parameter configurations of each layer in our ConvNet.}
\label{tab:ConvNet_configuration}
\end{table}

\paragraph{Direction.} We define 12 directions for each grid in Figure \ref{fig:ConvNet_structure} and use lookup table \(R^{12*200}\)to map each direction into 200-dimension embedding, which will be updated during model training. In order to tweak the CNN image embedding, we conduct element-wise multiplication between direction embedding and ConvNet embedding, resulting in a final 200-dimension image representation.

\paragraph{LINE.} We construct grid network and apply network embedding to capture the spatial correlation between neighboring grids. Specifically, each grid is a node in a network and the neighborhood relationship is converted into edge connection, while the weight on each edge is the reciprocal of Manhattan distance between two grids. In order to simplify the network, we consider at most 5-hop neighbors, which means any two grids that are more than 5 hops away will be disconnected. The spatial locality is actually implied by the network structure, which is represented by LINE \cite{tang2015line} in our work. In addition, we use 100-dimension for this structure representation and keep updated during model training.

\paragraph{Flow.} The same region may have different traffic conditions as time changes, therefore we need time-varying representation for each grid. To this end, we average the number of vehicle per grid per hour. Each grid is associated with a flow vector with 24 elements, representing the change of traffic conditions every hour.
Taking 100 vehicles as minimum unit, we have flow embedding as \(R^{1000 * 50}\). 

\paragraph{Attributes.} We consider three attribute information to improve the grid image representation: start time of the trip, vehicle ID and weather. In detail, we have three embedding lookup tables to represent each attribute: (i) we take one minute as minimum unit, given 7 days per week, we have start time embedding as \(R^{10080 * 30}\); (ii) different drivers may have different driving habits, we encode the driver ID as \(R^{25000 * 10}\); (iii) the weather condition is also critical for travel time estimation, we encode all kinds of weather into \(R^{400 * 10}\).

Finally, we get 50-dimension attribute vector after concatenation. Combined with 200-dimension image representation, 100-dimension LINE representation and 50-dimension flow representation, we generate 400-dimension vector per grid.

\subsection{Multitask Prediction}
Following previous works \cite{wang2018will,zhang2018deeptravel}, we use Bi-LSTM \cite{hochreiter1997long} to model sequential dependency and adopt residual connected layers for non-linear mapping. As for the prediction layer, we designed a different multitask structure, where each task is to estimate the travel time from origin grid to the current grid, to leverage the valuable information of sub-paths. 

\paragraph{Multitask Loss Function.} Given a trip with L grids, $\{g_1, g_2,\ldots,g_{L}\}$, we consider not only the mean absolute percentage error (MAPE) of the whole path from $g_1$ to $g_L$, but also the MAPE of sub-paths from $g_1$ to $g_l$. To this end, the final loss $\mathcal{L}$ is defined as
\begin{equation}
\resizebox{.53\linewidth}{!}{$
    \mathcal{L} = \frac{1}{L-1}\sum_{l=2}^{L} \left (w_l \cdot \frac{|\hat{T}_l - T_l|}{T_l} \right)
    $}
\end{equation}
where $T_{l}=t_{l} - t_{1}$ denotes the travel time from grid $g_1$ to $g_l$ and $\hat{T}_{l}$ denotes its estimation; $w_l$ is the predefined weight, $w_l = 2l/(L^2 + L-2)$, where $1<l\leq L$ and $\sum_{l=2}^L w_l =1$. In this way, we emphasize the longer sub-trips, to make sure model put more effort on whole trip estimation.

\subsection{Travel Time Estimation}

\paragraph{Path-aware Estimation.}
Given a \textit{path-aware} query, we map its path into grid sequence with driving directions. We then feed the grid sequence and the associated attribute information into the well-trained DeepI2T, which outputs whole travel time prediction.

\paragraph{Path-blind Estimation.}
We design a neighboring strategy to tackle the \textit{path-blind} query. Given a query with departure time, origin and destination locations, we first find the trips with the same origin and destination grids in the historical trips, namely neighboring trips. We predict the neighboring trips' travel time at the same departure date and time as the queried trip. Finally, the travel time of queried trip is estimated by weighting the (estimated) travel time of neighboring trips with their $\ell_1$ distances.
\begin{equation}
\resizebox{.5\linewidth}{!}{$
    \hat{T}_{test} = \frac{1}{N_e}\sum_{i=1}^{N_e}{\frac{L_{test}}{L_i} \hat{T}_i}
    $}
\end{equation}
where $N_e$ denotes the number of neighboring trips in the training set having the same origin and destination grids as, or in the neighboring grids of, the queried trip; $L_{test}$ and $L_i$ refer to the $\ell_1$ distances of the testing and neighboring trips, respectively; $\hat{T}_{test}$ and $\hat{T}_i$ refer to the estimated travel time of the testing and neighboring trips, respectively. 

\section{Experiments}

\subsection{Data Description}
\begin{figure}[htb!]
\centering
\centerline{\includegraphics[width=0.80\linewidth]{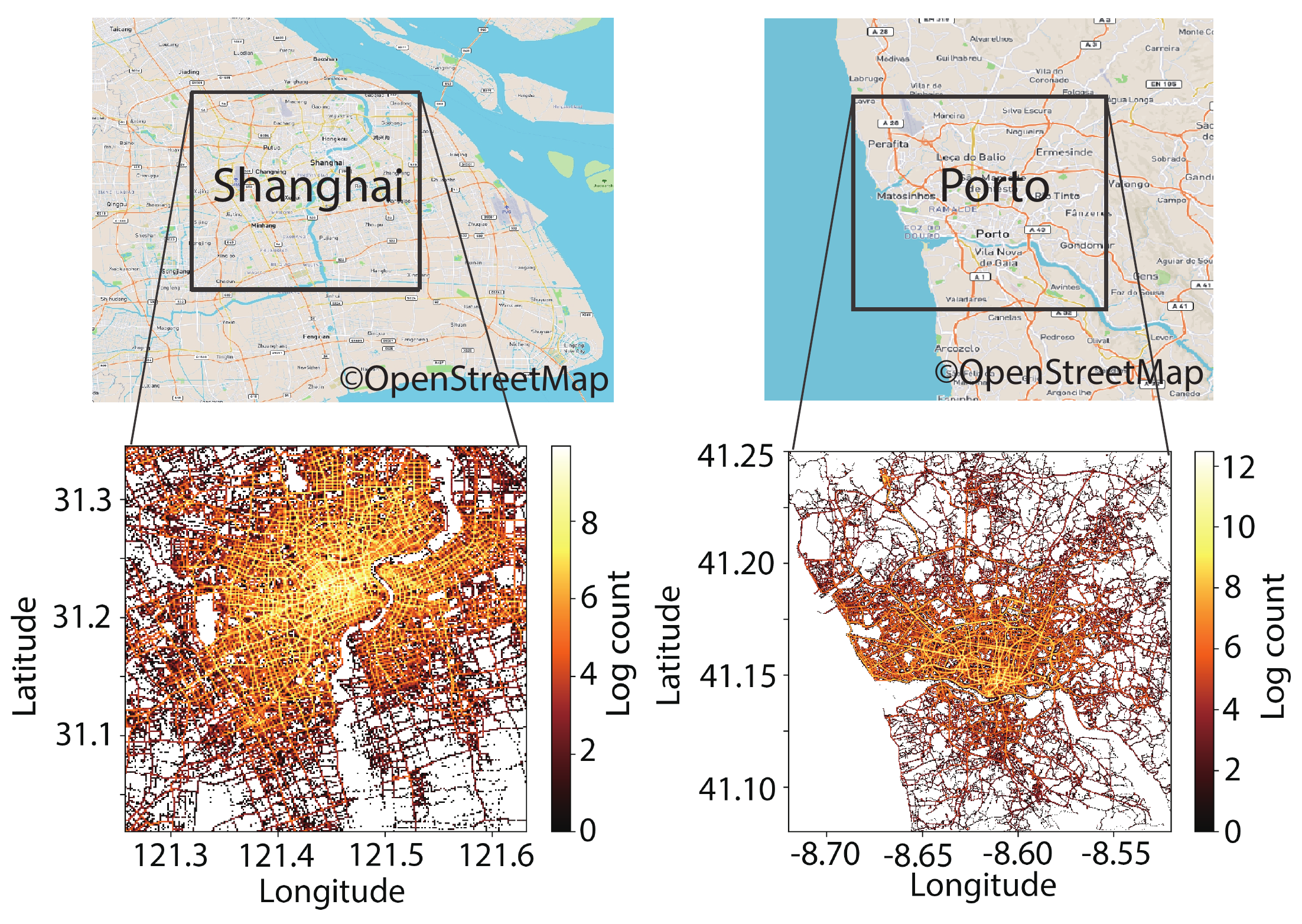}}
\caption{Study region and spatial distribution of GPS footprints in Shanghai in one week (left) and Porto in one year (right).}
\label{fig:map}
\end{figure}

In the experiments, large-scale datasets in two cities from different countries are adopted to validate the proposed DeepI2T. The study area and the distribution of raw GPS data are presented in Figure~\ref{fig:map}.
The statistical information of the datasets is shown in Table ~\ref{tab:datasets}. 

\paragraph{Shanghai Data.} Shanghai Data contains of the GPS trajectories of taxis from Apr. 1st to Jun. 15th in 2014, with sample interval ranges from 20s to 100s. We only keep the trips of taxis when they are transporting passengers. We select the data from Apr. 1st to May 31st for model training, and the remaining for testing. 

\paragraph{Porto Data.} The dataset is collected from 442 taxis running in the city of Porto, Portugal, from 1st Jul. 2013 to 30th Jun. 2014, and is publicly available~\cite{porto2018}. The measurement interval of the track points is fixed at 15s in the raw data. We select the data during the first 9 months for model training, and the remaining for testing. 


\begin{table}[htb!]
\begin{center}
\begin{tabular}{l|c|c}
\toprule
 \textbf{Dataset} & \textbf{Shanghai} & \textbf{Porto} \\
\midrule
area of study region & 1479.7$km^2$ & 419.0$km^2$ \\
\# grids (W$\times$H) & 200 $\times$ 175 & 100 $\times$ 85\\
size of grid & $\sim 200 m$ & $\sim 200 m$ \\ 
\# vehicles & $15,000$ & $442$ \\
\# trips per day & $170,000$ & $3,530$ \\
average sample rate & $55s$ & $15s$ \\
average travel time & $719s$ & $749s$ \\
travel time std & $469s$ & $593s$ \\
average travel distance & $4.53km$ & $5.79km$\\
\# days for training & 61 & 273\\
\# days for testing & 15 & 90\\
\bottomrule
\end{tabular}
\caption{The statistical information of the datasets}
\label{tab:datasets}
\end{center}
\end{table}

\subsection{Methods for Comparison}

\paragraph{Linear regression (LR).} Given the geographical positions of the origin and destination, we train the relation between the $\ell_1$ distance and travel time using a linear regression model.

\paragraph{Neighbor average (AVG).} For each trip in the training data sets, the average speed is calculated with the travel time and the $\ell_1$ distance between the origin and destination. During the testing phase, we estimate the travel speed by simply averaging the historical speeds of its neighbors, i.e., trips between the same origin and destination grids.

\paragraph{Temporally weighted neighbors (TEMP).}
Wang \textit{et al.} proposed a simple baseline method to estimate trip travel time by scaling the average speeds of its neighboring trips~\cite{wang2016simple}. The scaling factors are calculated by the temporal change of the average speeds of all trips in the city, i.e., relative temporal speed reference in~\cite{wang2016simple}. Comparing with AVG method, TEMP inspects the variance in average speeds of historical trips by their departure hours.

\paragraph{Gradient Boosting Machine (GBM).} Gradient boosting decision tree models have been used in travel time prediction~\cite{zhang2015gradient}. Several attributes of trips are fed into GBM model, including the departure time, the day in a week, the geographical coordinates of the origin and destination, and the $\ell_1$ distance, etc. The model is implemented using LightGBM~\cite{ke2017lightgbm}, configured with 500 trees with 1,000 leaves. 

\paragraph{DeepTTE.} It is a state-of-the-art end-to-end deep learning method for travel time estimation, learning the spatial and temporal dependencies of the raw GPS points in trajectories~\cite{wang2018will}. We use the code shared by authors to test our two datasets. However, we find DeepTTE tends to overfit when fed with uniformly sampled GPS trajectories (i.e., Porto data has fixed sample rate). In such case, DeepTTE simply learns the travel time by counting the number of points in one trace and fails in the testing. To fix it, we feed DeepTTE with the centroids of the grids traversed by trips for both training and testing data, as same as DeepI2T.

\paragraph{GridLSTM.} It is a degeneration of DeepI2T by removing ConvNet and direction embedding. It learns the travel time with the spatial and temporal relation of grids in one trace.

\subsection{Performance Evaluation}

We utilize three metrics to evaluate the performance of reference models, mean absolute error (MAE), mean absolute percentage error (MAPE), and satisfaction rate (SR). SR is defined as the fraction of trips which estimation error rates are not more than $10\%$ and higher SR indicates better performance. The formulas of these metrics are given as follow.
\begin{equation}
\resizebox{.55\linewidth}{!}{$
    MAE(T,\hat{T}) = \frac{1}{N}\sum_{i=1}^{N}{|T_i-\hat{T_i}|}
    $}
\end{equation}
\begin{equation}
\resizebox{.71\linewidth}{!}{$
    MAPE(T,\hat{T}) = \frac{1}{N}\sum_{i=1}^{N}{\frac{|T_i-\hat{T_i}|}{T_i}} \times 100\%
    $}
\end{equation}
\begin{equation}
\resizebox{.83\linewidth}{!}{$
    SR(T,\hat{T}) = \frac{1}{N} \sum_{i=1}^{N} \left ( \frac{|T_i-\hat{T}_i|}{T_i} \leq 10\% \right ) \times 100\%
    $}
\end{equation}
where $T_i$ and $\hat{T}_i$ are the actual and estimated travel time of the $i$th trip in the $N$ testing trips. 

In Table~\ref{tab:err_comp}, we present the estimation errors of reference methods. The first group of methods (LR, AVG, GBM and TEMP) are designed for \textit{path-blind} query and the second group of methods (DeepTTE and GridLSTM) are designed for \textit{path-aware} query. The best performance is highlighted per metric per group. As we can see, DeepI2T performs best in both groups in terms of MAPE. Compared with GridLSTM, the introducing of image representation promotes the MAPE of \textit{path-aware} queries by $10\%$ and $7\%$ in Shanghai and Porto, respectively. Regarding the \textit{path-blind} queries, DeepI2T shows competitive performance with TEMP in Shanghai, but evidently better performance in Porto.

\begin{table*}[!tb]
\begin{center}
\begin{tabular}{lllllllll}
\toprule
\multirow{2}{*}{\textbf{Method}} & & \multicolumn{3}{c}{\textbf{Shanghai}} & &  \multicolumn{3}{c}{\textbf{Porto}}
 \\ 
 \cmidrule{3-5} \cmidrule{7-9}
& & \textbf{MAE (s)} & \textbf{MAPE (\%)} &  \textbf{SR (\%)} & & \textbf{MAE (s)} & \textbf{MAPE (\%)} &  \textbf{SR (\%)}\\
\midrule
\textbf{LR} & & 186.5 & 27.64 & 23.87 & & 287.9 & 49.02 & 17.20\\
\textbf{AVG} & & 158.9 & 22.30 & 29.35 & & 235.86 & 30.43 & 24.65\\
\textbf{GBM} & & 144.3 & 22.55 & 30.45 & & 238.17 & 37.83 & 22.97 \\
\textbf{TEMP}~\cite{wang2016simple} & & \textbf{141.0} & 21.93 & \textbf{31.24} & & 231.10 & 29.84 & 25.67 \\
\hline
\rowcolor{Gray} \textbf{DeepTTE}~\cite{wang2018will} & & 147.61 & 19.02 & 31.13 & & 167.94 & 20.44 & 32.34 \\
\rowcolor{Gray} \textbf{GridLSTM} & & 117.12 & 16.98 & 37.00 & & 139.55 & 18.10 & 38.45 \\
\hline
 \textbf{DeepI2T} (\textit{path-blind}) & & 143.61 & \textbf{20.47} & 30.62 & & \textbf{186.65} & \textbf{25.28} & \textbf{30.08} \\
\rowcolor{Gray} \textbf{DeepI2T} (\textit{path-aware}) & & \textbf{105.43} & \textbf{15.20} & \textbf{42.23} & &  \textbf{128.26} & \textbf{17.08} & \textbf{38.97} \\
\bottomrule
\end{tabular}
\caption{Overall performance comparison on Shanghai and Porto Data. Path-aware methods are highlighted in gray shadow.}
\label{tab:err_comp}
\end{center}
\end{table*}

\begin{table}[htb]
\begin{center}
\begin{tabular}{lllllll}
\toprule
 \multirow{2}{*}{\textbf{Method}} & & 
 \multicolumn{2}{c}{\textbf{MAPE$_{peak}$ (\%)}} & & \multicolumn{2}{c}{\textbf{SR$_{peak}$ (\%)}} \\
\cmidrule{3-4} \cmidrule{6-7}
& & \textbf{AM} & \textbf{PM} & & \textbf{AM} & \textbf{PM}\\
\midrule
\textbf{LR} & & 27.89 & 26.17 & & 22.61 & 24.15\\
\textbf{AVG} & &  23.72 & 21.77 & & 26.39 & 28.30\\
\textbf{GBM} & & 23.74 & 24.13 & & 29.38 & 28.62 \\
\textbf{TEMP} & & 23.03 & 23.45 & & \textbf{30.23} & 29.44 \\
\hline
\rowcolor{Gray} \textbf{DeepTTE} & & 21.58 & 20.15 & & 26.72 & 29.08 \\
\rowcolor{Gray} \textbf{GridLSTM} & & 18.95 & 17.95 & & 33.40 & 34.62 \\
\hline
\makecell[l]{\textbf{DeepI2T}\\(\textit{path-blind})} & & \textbf{22.55} & \textbf{20.40} & & 27.29 & \textbf{29.52} \\
\rowcolor{Gray} \makecell[l]{\textbf{DeepI2T}\\(\textit{path-aware})} & & \textbf{16.89} & \textbf{15.75} & & \textbf{37.28} & \textbf{41.37} \\
\bottomrule
\end{tabular}
\caption{Peak hour performance comparison on Shanghai Data}
\label{tab:err_peak}
\end{center}
\end{table}

We also summarize the MAPE and SR of each model during the morning (7:00-9:00) and evening (16:00-18:00) peak hours in Shanghai in Table~\ref{tab:err_peak}. Comparing with the other two deep learning baselines, DeepI2T yields the best performance in terms of MAPE and SR on \textit{path-aware} queries during both peaks. Especially during the evening peak, the SR of DeepI2T reaches $41.37\%$, while DeepTTE and GridLSTM only can achieve $29.08\%$ and $34.62\%$, respectively. That indicates DeepI2T could provide acceptable time estimation for a large fraction of trip queries during traffic congestion. The \textit{path-blind} estimation also reaches competitive performance with the baselines during peak hours, except the SR during morning peak is weaker than TEMP.

\begin{figure}[tb!]
\centering
\centerline{\includegraphics[width=0.9\linewidth]{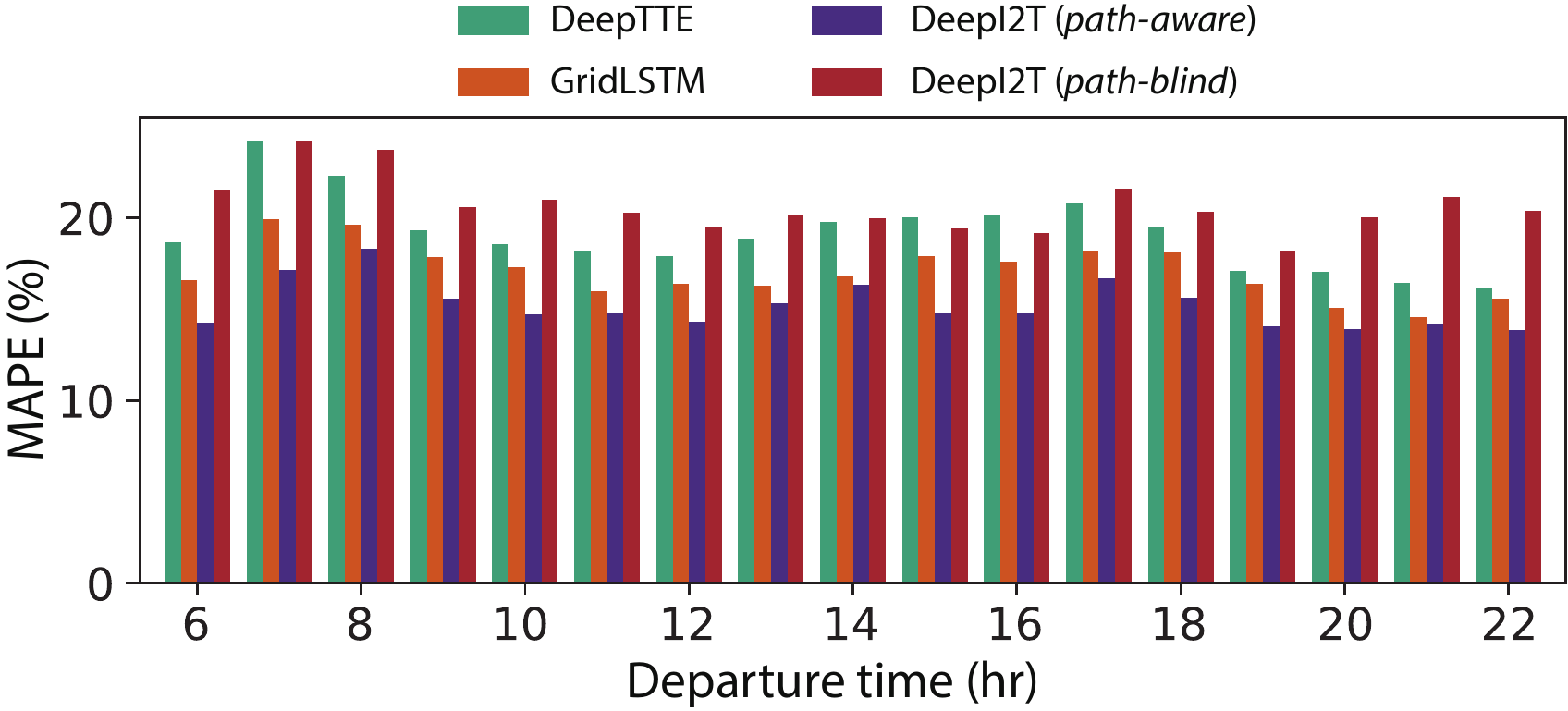}}
\caption{Estimation error for trips with different departure time.}
\label{fig:mape_hour}
\end{figure}

\begin{figure}[tb!]
\centering
\centerline{\includegraphics[width=0.96\linewidth]{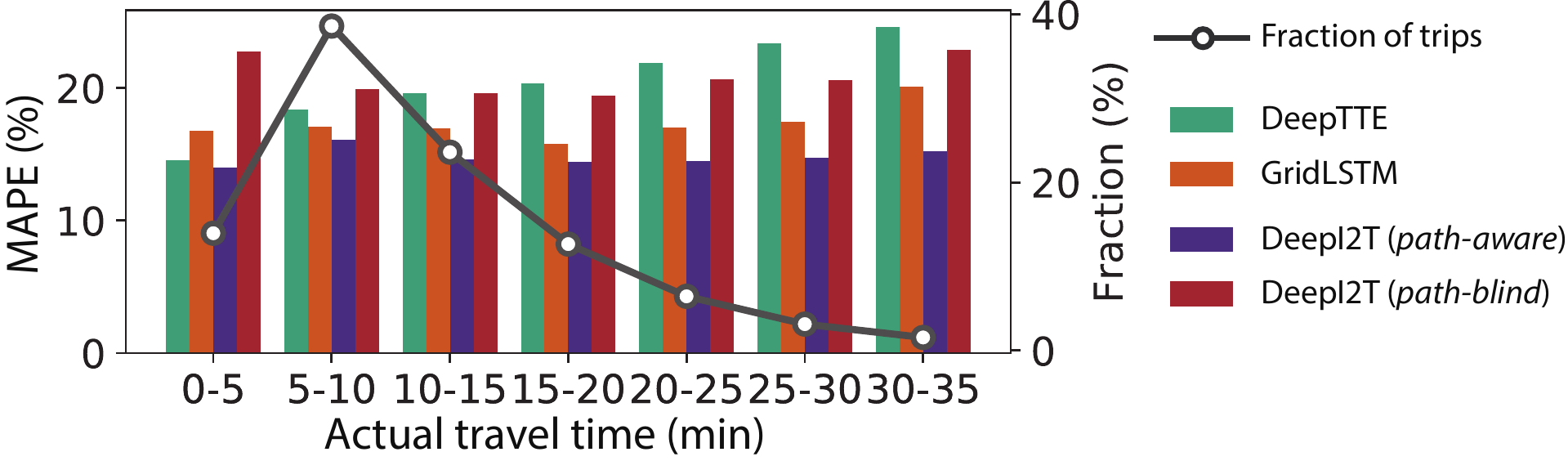}}
\caption{Estimation error for trips with different travel time. The right y-axis shows the distribution of testing trips.}
\label{fig:mape_time}
\end{figure}

\begin{figure}[tb!]
\centering
\centerline{\includegraphics[width=0.6\linewidth]{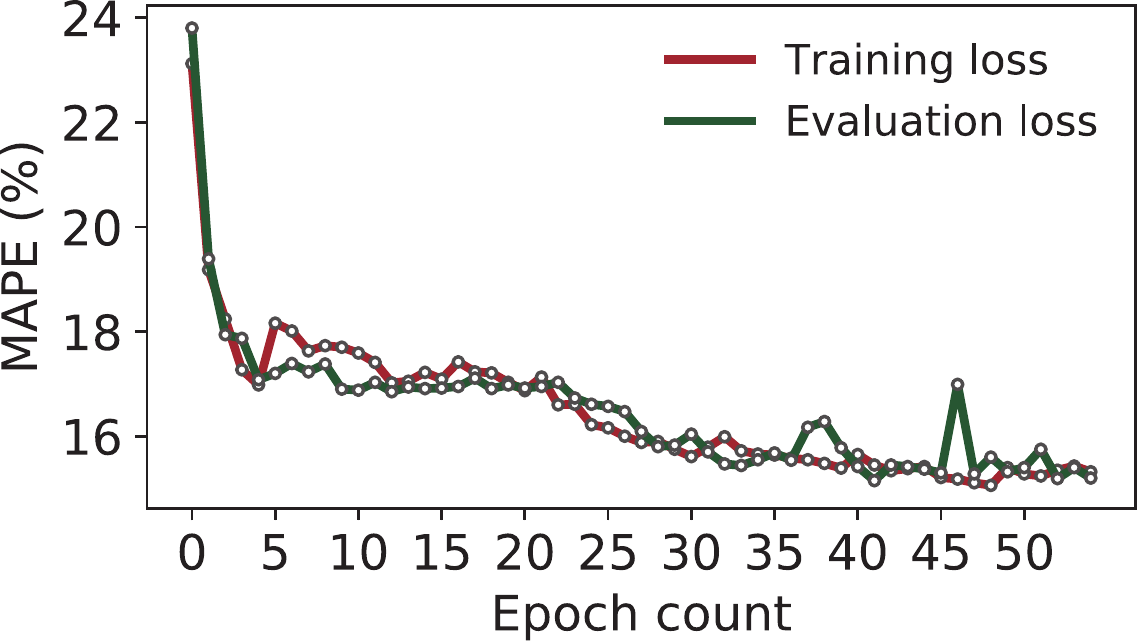}}
\caption{Evolution of estimation error during model training.}
\label{fig:mape_epoch}
\end{figure}

Further, we compare the performance of DeepTTE, GridLSTM and DeepI2T on Shanghai Data. Figure~\ref{fig:mape_hour} presents the estimation error per hour. All models show relatively weak performance during the peak hours due to the traffic congestion. Even so, DeepI2T generates better estimation than compared models during all the day. In Figure~\ref{fig:mape_time}, we compare the performance of these models per the actual travel time of queried trips. We find DeepTTE performs worse for longer trips, while the proposed DeepI2T has much stable performance for trips with varying travel time. In addition, we present the evolution of estimation error during the model training phase in Figure~\ref{fig:mape_epoch}. Overall, the MAPE decays stably along with the training time. The closeness between training and evaluation curves reflects that DeepI2T efficiently prevent overfitting during model training.

It's noteworthy that the performance of DeepTTE on our datasets is evidently weaker than the results in~\cite{wang2018will}in terms of MAPE. We argue the reasons are mainly in several aspects, (i) DeepTTE learns the dependency between two consecutive locations with their time gaps. It tends to fail when the sample rate of data is (nearly) constant; To fix this issue, we resample GPS points by using the centroids of grids. However, the time gap between two centroids may introduce some errors; (ii) DeepTTE is sensitive to the distance gap in the testing trip, which determines the similarity between testing and training data. How to select the best distance gap is not given in~\cite{wang2018will}; (iii) The trip lengths in~\cite{wang2018will} are approximately twice of ours on average. Normally, for a shorter trip, the MAPE tends to be higher than longer trip even they have similar MAE. Another state-of-the-art model is DeepTravel~\cite{zhang2018deeptravel}, which we didn't compare because the source code is not publicly available. We notice that the standard deviation of travel time for Porto Data in DeepTravel in~\cite{zhang2018deeptravel} is much smaller than ours ($348s$ vs. $593s$), indicating the task in our work is more challenging.

\section{Conclusions}
This work explores the potential of learning travel delay from urban layout images for a specific task, \textit{travel time estimation}. Our proposed DeepI2T model showed promising performance for both path-aware and path-blind scenarios on real-world datasets in Shanghai and Porto. Our attempt presents new opportunities of using publicly available morphological images to solve transportation problems.

\section*{Acknowledgments}

The work is supported by the Science and Technology Innovation Action Plan Project of Shanghai Science and Technology Commission under Grant No. 18511104202.


\clearpage

\bibliographystyle{named}
\bibliography{ijcai19}

\end{document}